\documentclass[fleqn,10pt]{olplainarticle}
\usepackage{url}

\title{EnviroLLM: Resource Tracking and Optimization for Local AI}
\author[1]{Troy Allen}
\affil[1]{Georgia Institute of Technology, College of Computing}

\begin{abstract}
Large language models (LLMs) are increasingly deployed locally for privacy and accessibility, yet users lack tools to measure their resource usage, environmental impact, and efficiency metrics. This paper presents EnviroLLM, an open-source toolkit for tracking, benchmarking, and optimizing performance and energy consumption when running LLMs on personal devices. The system provides real-time process monitoring, benchmarking across multiple platforms (Ollama, LM Studio, vLLM, and OpenAI-compatible APIs), persistent storage with visualizations for longitudinal performance analysis, and personalized model and optimization recommendations. The system includes LLM-as-judge evaluations alongside energy and speed metrics, enabling users to assess quality-efficiency tradeoffs when testing models with custom prompts.
\end{abstract}

\begin{document}

\flushbottom
\maketitle
\thispagestyle{empty}

\section*{Introduction}

The growing adoption of large language models (LLMs) in local, on-device settings has been driven by privacy concerns, company requirements, and reduced latency needs. However, this transition from centralized cloud deployments introduces new challenges in understanding, managing, and optimizing the performance and environmental impact of AI systems running on personal and company computing devices.

Current approaches to measuring AI's performance and environmental impact focus on training and inference costs in cloud-based settings at scale. This has left a significant gap in tools for individual users and smaller organizations to utilize in making decisions about model selection, optimization strategies, and sustainable AI practices.

This paper introduces EnviroLLM, an open-source toolkit designed to address this gap by providing resource tracking, benchmarking, and optimization recommendations for local LLM deployments. The system combines real-time process monitoring with persistent benchmark storage and interactive visualization, enabling users to compare models across custom tasks and track performance over time. This framework aims to contribute to sustainability as data center expansion accelerates and alternative methods for achieving the same results as cloud-based inference become increasingly important.

\section*{Related Work}

Research in AI sustainability and optimization typically focuses on training-phase and inference-phase energy consumption and performance in large-scale cloud deployment scenarios. Carbon estimation frameworks, including MLCO2 \citep{Lacoste2019}, GreenAlgorithm \citep{Lannelongue2020}, and CodeCarbon \citep{Fischer2025} were designed for data center operations. They rely on Thermal Design Power (TDP) approximations rather than real-time measurements during inference. These tools do not capture the resource patterns of personal devices seen when running large language models locally. Fischer et al. \citep{Fischer2025} demonstrated that CodeCarbon's estimation approaches can produce errors of up to 40\% when compared to ground-truth measurements, with TDP-based approximations particularly problematic in environments where dynamic frequency scaling, power management, and thermal throttling significantly impact actual consumption. These conditions are common in personal laptops and workstations running local LLMs.

Cloud-focused research has also developed sophisticated prediction models like LLMCarbon \citep{Faiz2023}. It introduces a polynomial regression framework for training emission prediction across GPU clusters, while OpenCarbonEval proposes a throughput modeling framework to capture workload fluctuations in data centers. \cite{Yu2024} has used modeling through Little's Law to capture these changes in usage patterns dynamically.

Research on local deployments suggests potential for emission reduction, but is marked by a lack of integrated tooling. \cite{Khan2025} show that quantization techniques, when combined with local inference, can achieve carbon emissions of up to 45\% in common model architectures like Llama, Phi, and Mistral. GPTQ, a post-training quantization method, is mentioned as being effective. However, there is a lack of studies focused on optimization techniques regarding predictive and pre-training configuration decisions.

This cloud-centric and post-training focus leaves a critical gap as the popularity of local LLMs continues to rise. Research has established that substantial emission reduction is possible in large-scale cloud settings, and EnviroLLM aims to solidify the notion that such reductions are also possible in smaller local settings.

Traditional evaluation metrics for text generation tasks often fail to capture nuanced aspects of response quality \citep{Zheng2023}. Recent work has explored using LLMs themselves as evaluators, with \citet{Zheng2023} introducing MT-Bench and demonstrating that GPT-4 as a judge achieves over 80\% agreement with human preferences. \citet{Ho2025} further validated this approach for extractive QA tasks, showing high correlation (0.85) with human judgments while identifying that performance varies by answer type. This body of work establishes LLM-as-a-judge as a viable alternative to expensive human evaluation, particularly relevant for benchmarking systems like EnviroLLM where users need to assess both efficiency and quality.

\section*{Methodology}

Running large language models (LLMs) on personal devices introduces challenges in measuring real-world performance and energy consumption. EnviroLLM addresses these challenges through a three-component architecture designed for local AI deployment analysis and optimization.

\subsection{System Architecture}

The framework consists of three core components that work together to provide insight into local AI resource consumption and potential optimizations:

1. \textbf{Command-Line Interface (CLI):} A lightweight tool that tracks LLM processes across platforms including Ollama, LM Studio, and llama.cpp. It allows for benchmarking across custom prompts and storage of the results. The CLI can also capture system performance metrics in real-time, including CPU usage, memory consumption, and GPU utilization.

2. \textbf{Web Dashboard:} Built with Next.js and React, this interface provides interactive visualizations for performance analysis. The dashboard displays system metrics, hardware-specific optimization recommendations, and grouped benchmark views with comparative charts for energy efficiency, inference speed, and quality metrics.

3. \textbf{Backend Processing Engine:} Integrated with Ollama's REST API and OpenAI-compatible endpoints, the backend enables measurement of model performance across multiple platforms, quantization levels, and prompts. By combining baseline consumption with active usage metrics and custom prompts, it provides a nuanced view of resource utilization tailored to the user's needs.

\subsection{Model Evaluation and Efficiency}

Local AI deployment requires balancing performance and resource consumption. \citet{Khan2025} demonstrated that quantization techniques combined with local inference can reduce carbon emissions by up to 45\% in model architectures like Llama and Phi. EnviroLLM enables testing of multiple models and quantization levels across identical tasks, capturing metrics including CPU usage, memory consumption, and energy expenditure during both short benchmark runs and extended monitoring periods.

\subsection{Smart Model Selection}
Model selection extends beyond simple parameter count. \citet{Haase2025} demonstrate that smaller models like Gemma-3 and Phi-4 can achieve competitive performance on common tasks. Research shows that quantization techniques can maintain accuracy while significantly reducing resource requirements \citep{Vijay2025}, informing EnviroLLM's approach to model evaluation. Users can create custom prompts or use EnviroLLM presets to get more realistic tests of their chosen model for their exact use case. EnviroLLM builds on the notion that model selection itself represents a critical optimization decision. This is particularly relevant in specialized domains like medical applications and code generation, where targeted models \citep{Vrettos2025, Vartziotis2025} may outperform general-purpose larger models on specific tasks.

\subsection{Benchmarking Made Simple}

The benchmarking workflow consists of five steps:

1. Select models through CLI or web interface

2. Run identical tests across all chosen variants

3. Automatically store results with prompt-based grouping for comparison

4. View interactive visualizations of benchmark history and performance trends

5. Export data for external analysis or reset baselines as needed

A single command like \texttt{envirollm benchmark --models llama3:8b,phi3:mini} provides instant, detailed model comparisons for Ollama deployments, while \texttt{envirollm benchmark-openai --url http://localhost:1234/v1 --model llama-3-8b} enables benchmarking of LM Studio, vLLM, and other OpenAI-compatible APIs. This approach builds on \cite{Yu2024}'s work in understanding dynamic usage patterns for local AI optimization.

\section*{Implementation and Features}

EnviroLLM is distributed via npm. The CLI detects running LLM processes across major frameworks including Ollama, LM Studio, llama.cpp, Text Generation WebUI, KoboldCPP, and vLLM. The system samples CPU usage, memory consumption, GPU utilization, and power consumption at two-second intervals.

\subsection{Measurement Infrastructure}

EnviroLLM's resource tracking relies on several system-level APIs for monitoring. CPU and memory metrics are captured via Python's \texttt{psutil} library, which provides process-specific resource consumption data at
two-second intervals. For NVIDIA GPUs, the system uses \texttt{pynvml} (NVIDIA Management Library) to query GPU utilization, memory usage, temperature, and real-time power draw. When GPU power telemetry is unavailable, the system estimates power consumption using a baseline system draw combined with CPU and GPU utilization percentages. Energy consumption (Wh) is calculated by integrating power measurements over the inference duration. The CLI automatically detects running LLM processes by matching against known process names (ollama, lmstudio, llama-server, etc.) and monitors their resource footprint in real-time.

The CLI provides automated benchmarking for multiple platforms. The \texttt{envirollm benchmark} command supports Ollama models, while \texttt{envirollm benchmark-openai} enables benchmarking of LM Studio, vLLM, text-generation-webui, and other OpenAI-compatible APIs. Both commands support custom prompts, allowing users to test models on domain-specific tasks. The system executes inference while capturing token-level metrics including generation speed, energy consumption, and response tokens. Quantization levels are automatically detected from model names, and actual LLM responses are stored for quality comparison across configurations.

The web dashboard provides three views: a monitoring interface showing real-time system metrics, an optimization page with hardware-specific recommendations, and a benchmarking interface with unified model comparison across platforms. The benchmarking interface features tabs for Ollama (multi-model selection), LM Studio (direct integration), and custom OpenAI-compatible APIs (vLLM, text-generation-webui). Users can configure custom prompts and view comparative results including energy consumption, inference speed, and response quality.  

\subsection{Data Management}
EnviroLLM uses SQLite-based storage to enable tracking across benchmark runs. The database stores complete results—model configurations, quantization levels, energy metrics, inference speeds, and LLM responses—addressing a critical gap in existing tools: the ability to track performance over time. Each benchmark is indexed by a \texttt{prompt\_hash}, enabling comparison of models tested with identical prompts. This provides three key capabilities:

\begin{itemize}
\item \textbf{Automatic Storage:} Results persist, building a record of model performance on the user's specific hardware.
\item \textbf{CSV Export:} The \texttt{envirollm export} command enables data analysis in external tools or integration with outside workflows.
\item \textbf{Data Management:} The \texttt{envirollm clean} command removes stored benchmarks when users need to reset their baseline.
\end{itemize}

Benchmark results also include timestamps displayed in the web interface to enable chronological tracking of model performance. This allows users to identify performance regressions, validate optimization improvements, and correlate resource consumption patterns with specific testing periods. This transforms EnviroLLM from a one-time measurement tool into a platform for understanding trends and detecting performance regressions in local AI deployments.

\subsection{Grouped Benchmark Visualizations}

Traditional benchmark tools report isolated measurements, making cross-model comparison difficult. EnviroLLM addresses this through prompt-grouped analysis and visualizations. The web dashboard clusters results sharing the same \texttt{prompt\_hash}, enabling comparison of how different models respond to identical tasks. This grouped analysis is particularly valuable for domain-specific deployments. Users can create custom prompts reflecting their actual workload, benchmark multiple candidates, and identify which configuration offers the best balance of accuracy, speed, and efficiency.

\section*{Use Cases and Applications}

EnviroLLM addresses practical scenarios where resource awareness influences deployment decisions. The toolkit operates in two modes: a standalone CLI for monitoring running processes, and a web-based frontend with benchmarking and visualization capabilities.

\textbf{Process Monitoring:} Users running LLMs through Ollama, LM Studio, llama.cpp, or other frameworks can use the CLI to monitor resource consumption of their active inference workloads. The system automatically detects running model processes and tracks CPU usage, memory consumption, GPU utilization, and power draw in real-time. This enables users to understand the actual resource footprint of their day-to-day LLM usage without requiring separate benchmark runs.

\textbf{Model Comparison and Selection:} Users evaluating different models can use the benchmarking interface to compare multiple candidates on identical prompts. For example, running \texttt{envirollm benchmark --models llama3:8b,phi3:mini,gemma2:9b} tests each model on the same tasks and stores results for comparison. The web dashboard displays benchmark history with interactive visualizations, allowing users to identify which models offer the best balance of speed, efficiency, and response quality for their specific use cases.

\textbf{Organizational Deployment Decisions:} Small organizations deploying LLMs for internal use benefit from both the optimization recommendations and comparative benchmarking. The system analyzes available RAM and GPU memory to suggest appropriate model sizes, while custom prompts allow teams to evaluate models on domain-specific tasks. The persistent benchmark history enables teams to document their model selection rationale with measured performance data.

\textbf{Performance Regression Detection:} Privacy-focused applications requiring local inference can track resource usage over time using the benchmark history feature. When updating models or switching between quantization levels, users can compare current performance against historical baselines to identify efficiency regressions. The web dashboard's visualization interface makes these trends immediately apparent, while the \texttt{envirollm clean} command allows resetting stored data when needed.

\textbf{Sustainability Research and Education:} Students and researchers studying AI sustainability can use EnviroLLM to gather reproducible data on energy consumption patterns. The CLI's real-time monitoring captures actual usage during specific tasks, while the benchmarking system enables controlled comparisons across models and platforms. Multi-platform support (Ollama, LM Studio, vLLM, OpenAI-compatible APIs) allows comparative studies that reveal how inference framework choice impacts resource consumption alongside model selection decisions.

\section*{Results}

Benchmarks were conducted on a single workstation (Intel Core i7, 32GB RAM, NVIDIA GTX 3080, Windows 11) to validate EnviroLLM's measurement capabilities across diverse workloads. Three models were evaluated: \texttt{gemma3:1b} (Ollama), \texttt{gemma-3-1b} (LM Studio), and \texttt{gemma-3n-e4b} (LM Studio). Each model was tested on five prompts representing different task categories: explanation (quantum computing), code generation (bubble sort), summarization (machine learning concepts), long-form generation (Tokyo itinerary), and analytical writing (renewable energy analysis).

\subsection*{Quality Evaluation}

Quality scores were generated using a hybrid evaluation approach. When Ollama was available, we employed the LLM-as-a-judge methodology \citep{Zheng2023} using \texttt{gemma3:1b} as the judge model. The evaluation prompt instructed the judge to rate responses on a 0-100 scale based on four criteria: accuracy (factual correctness), completeness (coverage of key points), clarity (ease of understanding), and relevance (staying on topic). 

When LLM-as-a-judge was unavailable, the system fell back to heuristic scoring based on textual features: completeness, vocabulary diversity, length appropriateness, and structure. This hybrid approach ensured consistent quality measurement across benchmark runs while maintaining the reliability benefits of LLM-based evaluation when available \citep{Ho2025}. All quality scores reported in this paper were generated using the LLM-as-a-judge method with \texttt{gemma3:1b} running locally via Ollama.

\subsection*{Cross-Task Performance Analysis}

Table~\ref{tab:cross-task} shows performance variation across task types. Energy consumption and generation speed varied significantly based on prompt complexity and output length, with short summarization tasks consuming as little as 0.087 Wh, while long-form generation required up to 3.83 Wh.

\begin{table}[h]
\centering
\small
\begin{tabular}{llrrrrr}
\hline
Model & Task Type & Energy (Wh) & Speed (tok/s) & Wh/tok & Quality & Tokens \\
\hline
\multicolumn{7}{l}{\textbf{gemma-3-1b (LM Studio, Q4)}} \\
 & Explanation & 0.410 & 110.3 & 0.000565 & 95 & 725 \\
 & Code Gen & 0.348 & 178.2 & 0.000396 & 95 & 877 \\
 & Summarization & 0.087 & 141.1 & 0.000522 & 95 & 167 \\
 & Long-form & 0.493 & 197.6 & 0.000354 & 75 & 1393 \\
 & Analysis & 0.450 & 172.7 & 0.000468 & 75 & 961 \\
\hline
\multicolumn{7}{l}{\textbf{gemma3:1b (Ollama, Q4)}} \\
 & Explanation & 0.457 & 121.9 & 0.000583 & 75 & 784 \\
 & Code Gen & 0.419 & 186.8 & 0.000365 & 95 & 1149 \\
 & Summarization & 0.113 & 124.2 & 0.000575 & 95 & 196 \\
 & Long-form & 0.520 & 193.7 & 0.000383 & 75 & 1356 \\
 & Analysis & 0.510 & 189.1 & 0.000392 & 95 & 1302 \\
\hline
\multicolumn{7}{l}{\textbf{gemma-3n-e4b (LM Studio, Q4)}} \\
 & Explanation & 1.115 & 43.2 & 0.001697 & 75 & 657 \\
 & Code Gen & 2.141 & 43.5 & 0.001810 & 95 & 1183 \\
 & Summarization & 0.371 & 39.8 & 0.001863 & 95 & 199 \\
 & Long-form & 3.830 & 42.3 & 0.001862 & 95 & 2057 \\
 & Analysis & 2.257 & 42.6 & 0.001711 & 75 & 1319 \\
\hline
\end{tabular}
\caption{Performance across task types showing energy consumption, speed, and quality scores for three model configurations. Quality scores (0-100) were generated using LLM-as-judge evaluation.}
\label{tab:cross-task}
\end{table}

\textbf{Key Findings:}
\begin{itemize}
\item \textbf{Task complexity drives resource consumption:} Summarization tasks (167-199 tokens) consumed 0.087-0.371 Wh, while long-form generation (1356-2057 tokens) required 0.493-3.83 Wh—up to 44× more energy for the larger model.

\item \textbf{Energy per token reveals efficiency:} Despite higher absolute energy use on longer tasks, the \texttt{gemma-3-1b} model maintained consistent efficiency (0.000354-0.000583 Wh/token), while \texttt{gemma-3n-e4b} consumed 3-5× more energy per token (0.001697-0.001863 Wh/token).

\item \textbf{Quality-efficiency tradeoffs vary by task:} Both smaller models (\texttt{gemma-3-1b}, \texttt{gemma3:1b}) achieved 95/100 quality on code generation and summarization while consuming significantly less energy than the larger model.
\end{itemize}

\subsection*{Platform Comparison: Ollama vs LM Studio}

Table~\ref{tab:platform} compares the same base model (\texttt{gemma-3-1b}, Q4 quantization) across Ollama and LM Studio platforms on the "Explain quantum computing" prompt.

\begin{table}[h]
\centering
\begin{tabular}{lrrrrr}
\hline
Platform & Energy (Wh) & Duration (s) & Speed (tok/s) & Wh/token & Quality \\
\hline
Ollama (gemma3:1b) & 0.457 & 6.43 & 121.9 & 0.000583 & 75 \\
LM Studio (gemma-3-1b) & 0.410 & 6.57 & 110.3 & 0.000565 & 95 \\
\hline
\end{tabular}
\caption{Platform comparison for the same model architecture and quantization level. Both platforms show comparable resource consumption, with variation primarily in quality scores.}
\label{tab:platform}
\end{table}

Platform overhead is minimal when running identical models—both consumed approximately 0.4-0.5 Wh with similar speeds (110-122 tok/s). The 20-point quality difference (75 vs 95) reflects variation in the actual model responses
rather than platform overhead, as the energy and speed metrics remained nearly identical. This indicates that platform choice may influence model behavior (possibly through different default parameters or inference optimizations) while having minimal impact on computational efficiency.

\subsection*{Model Architecture Impact}

Table~\ref{tab:architecture} compares two model architectures at the same quantization level running on the same platform (LM Studio).

\begin{table}[h]
\centering
\begin{tabular}{lrrrrr}
\hline
Model & Energy (Wh) & Speed (tok/s) & Wh/token & Avg Quality & Token Range \\
\hline
gemma-3-1b & 0.358 & 160.0 & 0.000460 & 87 & 167-1393 \\
gemma-3n-e4b & 1.943 & 42.3 & 0.001789 & 87 & 199-2057 \\
\hline
\textbf{Ratio} & \textbf{5.4×} & \textbf{3.8×} & \textbf{3.9×} & \textbf{1.0×} & — \\
\hline
\end{tabular}
\caption{Architecture comparison showing averaged metrics across all five tasks. The larger model consumes 5.4× more energy while producing similar quality scores.}
\label{tab:architecture}
\end{table}

The \texttt{gemma-3n-e4b} model consumed 5.4× more total energy and 3.9× more energy per token than \texttt{gemma-3-1b}, while running 3.8× slower. Despite this substantial resource difference, both models achieved identical average quality scores (87/100), demonstrating that model architecture dominates resource consumption independent of task-specific performance.

\subsection*{Quality-Efficiency Tradeoffs}

Our quality scores (Table~\ref{tab:cross-task}) were generated using LLM-as-judge evaluation, a methodology shown to correlate highly with human judgment \citep{Zheng2023, Ho2025}. Consistent with findings from \citet{Ho2025}, we observe that quality scores vary by task type and answer complexity. For instance, both smaller models (\texttt{gemma-3-1b}, \texttt{gemma3:1b}) achieved 95/100 quality on code generation and summarization tasks while consuming significantly less energy than larger models.

This demonstrates a key insight for local deployment: users can often achieve high-quality results with smaller, more efficient models, particularly for well-defined tasks. However, for more complex reasoning tasks, the quality-efficiency tradeoff may favor larger models despite higher energy consumption.

\subsection*{Measurement Consistency}

To validate EnviroLLM's measurement reliability, we examined variance across tasks for the same model configuration. The \texttt{gemma-3-1b} model showed consistent energy efficiency across diverse workloads:
\begin{itemize}
\item Energy per token: 0.000354-0.000565 Wh/token (60\% range)
\item Speed: 110.3-197.6 tok/s (79\% variation)
\item The variation correlates with task characteristics rather than measurement error: code generation tasks (requiring structured output) showed faster speeds but maintained consistent per-token energy consumption.
\end{itemize}

These results demonstrate that EnviroLLM provides reliable comparative benchmarks across platforms, models, and task types, enabling users to make data-driven decisions about local LLM deployment.

\section*{Software Availability}

EnviroLLM is available as open-source software at \url{https://github.com/troycallen/enviro-llm} under the MIT License. The CLI tool is distributed via npm as \texttt{envirollm} and can be run without installation using \texttt{npx envirollm} or installed globally with \texttt{npm install -g envirollm}. The web dashboard is accessible at \url{https://envirollm.com}. The system requires Node.js 14+ and Python 3.7+ as dependencies.

\section*{Limitations}

While EnviroLLM provides reliable comparative benchmarks, several limitations should be noted. First, all measurements were conducted on a single hardware configuration (Intel Core i7, NVIDIA GTX 3080, Windows 11), which may not generalize to other system architectures, particularly Apple Silicon or AMD GPUs. Second, our validation focused on Gemma model variants; broader model families (Llama, Mistral, Qwen) may exhibit different efficiency characteristics. Third, we selected \texttt{gemma3:1b} as the judge model to balance evaluation quality with resource efficiency, enabling rapid assessment without substantial computational overhead. However, \citet{Zheng2023} demonstrated that stronger models like GPT-4 achieve higher agreement with human preferences (80\%). Future work could validate our quality scores against human judgments or stronger judge models to quantify the accuracy-efficiency tradeoff more accurately. Finally, our energy measurements capture GPU and CPU consumption but cannot account for all running processes on a local machine. This may have led to some inaccuracies despite attempted controls.

\section*{Future Work}

While EnviroLLM provides a solid foundation for tracking resources and exploring options with local LLMs, several directions could enhance its utility for users and organizations.

\subsection{Batch Benchmarking}

The current implementation of EnviroLLM requires manual model selection or command-line input for each individual benchmark run. Future work should focus on introducing batch benchmarking, allowing users to test multiple models across a variety of tasks simultaneously. This would transform EnviroLLM into a systematic evaluation suite useful not only to users, but entire organizations. Combining this with integration into existing CI/CD pipelines would allow for benchmarking and comparisons against established baselines.

\subsection{Improved Optimization Recommendations}

EnviroLLM provides generic hardware-based recommendations based on available RAM and GPU specifications in its current state. An improved recommendation system based on the user's historical benchmarking data, hardware specifications, and desired task could recommend specific model-quantization pairings most tailored to the user's needs. This could enable users to make informed decisions without the need for exhaustive benchmarking over long periods of time.

\subsection{Optimization Loop}

A final useful direction for future work is adding a real-time optimization loop that selects models and model settings based on the task being processed. EnviroLLM could run small, fast tests to estimate speed, energy use, and quality for several model and quantization options. It would then choose the configuration that meets a user’s quality needs at the lowest cost. This would make the system adaptive during regular use and provide increased efficiency for applications where prompt types vary over time.

\section*{Acknowledgments}

I'd like to thank Dr. Nicholas Lytle and Dante Ciolfi for their guidance on this project and opportunity to work under the Computing Research Opportunities for Conservation and Sustainability (CROCS) lablet at Georgia Tech.

\section*{Conclusion}

EnviroLLM provides practical tools for measuring and optimizing local LLM deployments. By enabling direct comparison of models, quantization levels, and platforms across different architectures and prompts, the toolkit helps users make informed decisions about local AI. The system demonstrates that model architecture choices have greater impact on efficiency than platform selection, and that measurement consistency across runs validates the reliability of local benchmarking approaches. As local AI deployment grows, tools like EnviroLLM become essential for understanding the environmental and performance tradeoffs in on-device inference.

\nocite{*}
\bibliography{envirollm}

\end{document}